\documentclass{article}
\PassOptionsToPackage{numbers}{natbib}

\usepackage{optml}
\usepackage{amsmath}
\usepackage[T1]{fontenc}
\usepackage{hyperref}
\usepackage{url} 
\usepackage{booktabs}
\usepackage{amsfonts}
\usepackage{nicefrac}
\usepackage{microtype}
\usepackage{graphicx}

\title{Using stochastic computation graphs formalism\\ for optimization of sequence-to-sequence model}

\author{\name Eugene Golikov \email{golikov.e.a000@gmail.com}\\
  \name Vlad Zhukov \email{vladzhukovtirko@gmail.com}\\
  \name Maksim Kretov \email{kretovmk@gmail.com}\\
  \addr{Moscow Institute of Physics and Technology, Neural Networks and Deep Learning Lab}\\
  }

\begin{document}
\maketitle

\begin{abstract}
Variety of machine learning problems can be formulated as an optimization task for some (surrogate) loss function. Calculation of loss function can be viewed in terms of stochastic computation graphs (SCG). We use this formalism to analyze a problem of optimization of famous sequence-to-sequence model with attention and propose reformulation of the task. Examples are given for machine translation (MT). Our work provides a unified view on different optimization approaches for sequence-to-sequence models and could help researchers in developing new network architectures with embedded stochastic nodes. 
\end{abstract}

\section{Introduction}
\label{intro}

Stochastic computation graph is a directed acyclic graph which includes both deterministic nodes and conditional probability distributions \cite{scg}. Leaves of the computation graph are cost nodes whose sum can be attributed to total loss function of a machine learning model. By taking the expectation of sum of cost nodes w.r.t. random variables we receive expected loss. Having defined loss function as mathematical expectation of costs calculated by SCG, we can view losses for many seemingly different problems in a uniform way:

\begin{equation} \label{eq:scg_loss}
L(x, y_{true}, \theta,\theta') = \mathbb{E}_{z \sim p(z;\theta)}\sum_{c \in \mathcal{C}} c(x, y_{true}, z, \theta')
\end{equation}

Here $x$ are inputs, $y_{true}$ are corresponding ground-truth outputs, $z$ are random variables with probability distributions parameterized by $\theta$, $\theta'$ are other parameters of a model. $\mathcal{C}$ is a set of cost nodes, $c$. We refer reader to the seminal article \cite{scg} where authors provide examples of such reformulation for reinforcement learning (RL) problem setup and generative models.

Once we casted loss function as an expectation of costs calculated by SCG, we need to calculate its derivatives in order to use effective gradient-based approaches for optimization. Given that parameters of the model may be included in both probability distributions and deterministic non-linear transformations, the task of calculating the gradient may become problematic:

\begin{enumerate}
\item Closed analytical formula for gradient usually cannot be derived, and even formulae that include expectations are cumbersome.
\item If computation of gradient involves taking integrals numerically, samples often show large variance.
\end{enumerate}

As for the first problem, there is a general formula \cite{scg} which can be considered as a generalization of the well-known REINFORCE rule \cite{reinforce} and the usual backpropagation algorithm for deterministic computation graphs. It provides a straightforward way to obtain an unbiased estimate of the gradient of loss function by approximating expectations with Monte-Carlo samples. This is where the second problem comes to play. Naive numerical calculation of gradient may fail because of large variance of samples and some additional tricks are needed to make the procedure more sample-efficient: control variates \cite{var_red}, common random numbers etc. 

Recently, reparameterization trick for continuous distributions \cite{vae} has become popular for the same purposes and a similar procedure \cite{gumbel, concrete} was developed for discrete distributions. In the latter case, we introduce bias in gradient estimate but decrease the variance.

Stochastic computation graphs provide a convenient framework for the analysis of machine learning models. It also encourages us to use general variance reduction techniques for Monte-Carlo integration, not restricting ourselves to some "practices" which are common in fields where similar problems occur (such as baselines for lowering variance of score function estimator in RL).

The contributions of this work are as follows:

\begin{enumerate}
\item Re-formulation of an example NLP model using the SCG formalism, thus describing different optimization approaches in common terms.
\item Analysis of the existing training procedures for this model using the SCG formalism.
\item Testing different variance reduction techniques for the efficient optimization of the model.
\end{enumerate}

\section{Sequence-to-sequence model}
\label{seq2seq}

Sequence-to-sequence architectures (seq2seq) are a wide class of models that produce a sequence of tokens from an arbitrary input. Some examples of their applications are machine translation \cite{seq2seq}, summarization \cite{first_summ}. 

\begin{figure}[h]
  \centering
  \includegraphics[scale=0.46]{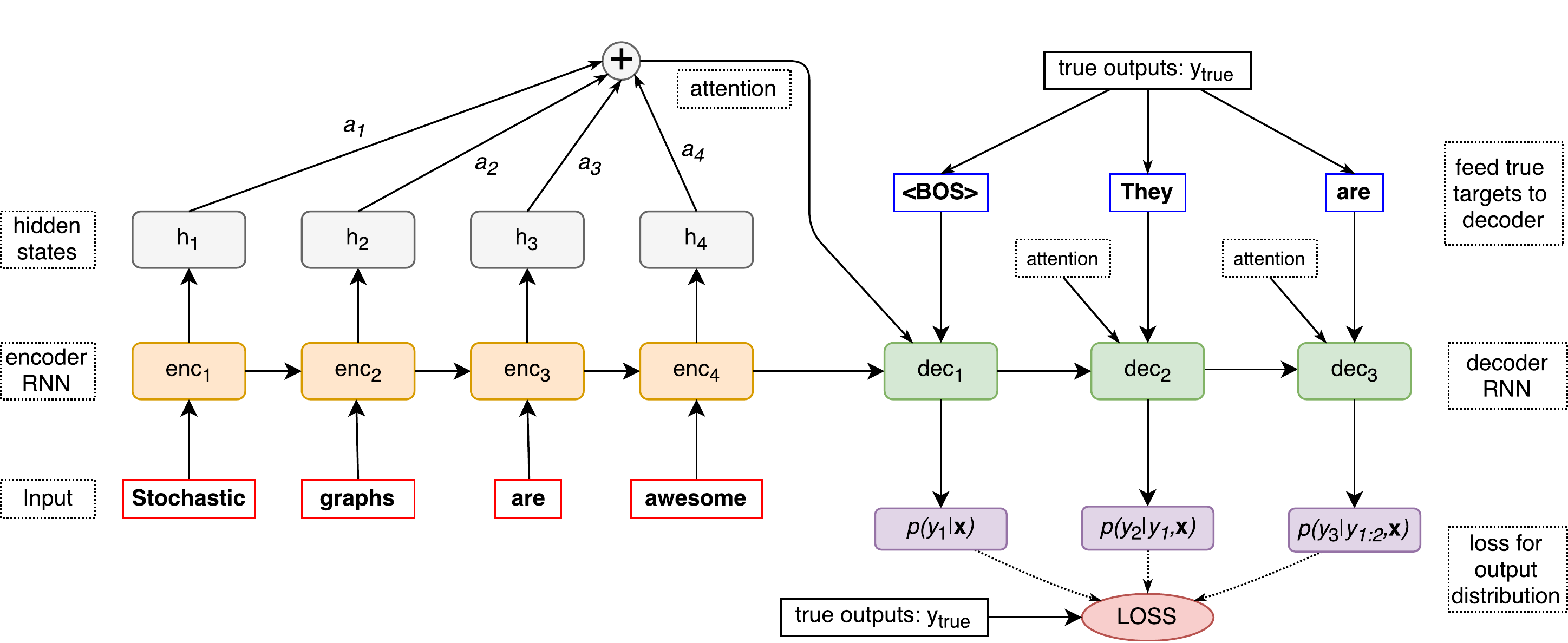}
  \caption{An example of sequence-to-sequence model with attention. Calculation of cross-entropy loss function with teacher forcing is shown. <BOS> is a token that denotes start of a sentence. All nodes in the graph are calculated deterministically and denoted by rounded rectangles.}
  \label{fig:seq2seq} 
\end{figure}

Widely used approach to training of seq2seq models is maximizing the likelihood of each successive target token conditioned on the input sequence and the history of target tokens. This approach is known as teacher forcing \cite{teacher_forcing} and is illustrated on Figure~\ref{fig:seq2seq}.  Although proved to be effective in practice, at test time it does not give model the access to correct tokens, but only to its own predictions. Hence inference procedure is different at training and test time. This leads to two major issues \cite{bso}:

\begin{enumerate}
  \item Exposure bias. Model is not exposed to its own outputs during training.
  \item Loss-evaluation mismatch. During training we optimize a differentiable metric, but measure quality with another metric (such as BLEU \cite{bleu}).
\end{enumerate}

Several approaches have been developed in order to mitigate these issues. 

\begin{figure}[h]
  \centering
  \includegraphics[scale=0.46]{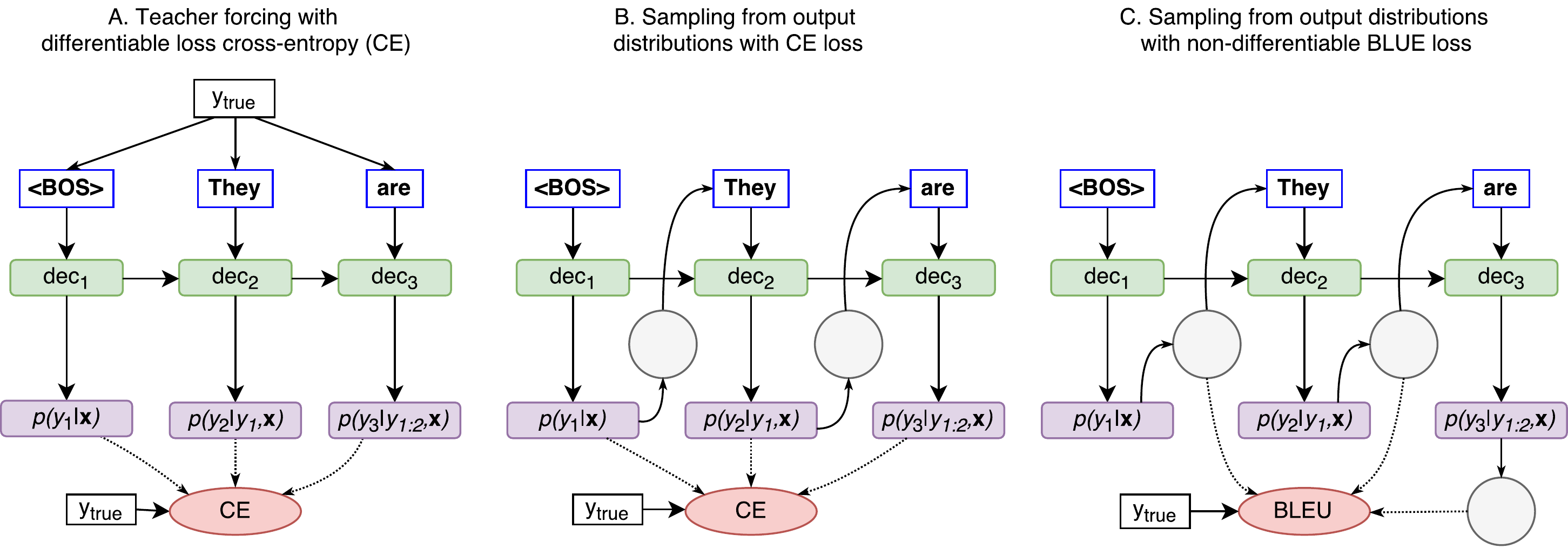}
  \caption{Different approaches to optimization of seq2seq model. For brevity, only decoder part is shown, encoder is the same as on Figure~\ref{fig:seq2seq}. Circles denote stochastic nodes where sampling is performed. A. Usual teacher forcing approach. B. Outputs from previous step are sampled and fed at the next time step. C. Same as in case B, but a non-differentiable metric is being optimized.}
  \label{fig:scg} 
\end{figure}

One is alternating regime of "scheduled sampling" \cite{scheduled_sampling} where model can receive either inputs from target sequences or samples from the output distribution at different training steps randomly. Log likelihood of correct labels is optimized. This method is developed within supervised learning paradigm (cases A and B in Figure~\ref{fig:scg}.

Second approach is to consider sampling from output distribution as an agent's action, thus considering the task in RL paradigm (case C in Figure~\ref{fig:scg}). Because of that, direct optimization of target non-differentiable metric can be performed using RL techniques \cite{abs_sumariz}.

In our opinion, there is nothing conceptually different between these approaches. They can be unified if we define loss function of seq2seq model in stochastic graph formalism as in Figure~\ref{fig:scg}. Then the expectation in equation ~\ref{eq:scg_loss} corresponds to sampled words from output distributions. We can consider these intermediate outputs as latent variables which are marginalized to get loss function depending only on external data and parameters of a model. Given general formula \cite{scg} for an unbiased estimate of gradient of corresponding loss, we now focus on its practical (i.e. data) efficiency. 

\section{Results and discussion}
\label{discussion}

We optimized all three losses shown in Figure~\ref{fig:scg} in common SCG framework for MT task\footnote{Our code is available on GitHub: \url{https://github.com/deepmipt/seq2seq_scg}}. The results are given in Table~\ref{results_mt} in the form "mean $\pm$ std". We performed 5 runs for each experiment in order to compute standard deviation. For the approach B ("feed samples"), when outputs from the previous step are sampled and fed at the next time step, we tested three different approaches to optimization. For all experiments we used early stopping (computation stops when loss on validation set stop decreasing, approximately between 10000-th and 20000-th batch), except for direct BLEU optimization, when we stopped after 100000 batches, not waiting for model to start overfitting. For direct optimization of BLEU (approach C on Figure~\ref{fig:scg}) control variates as a variance reduction technique were implemented.

The choice of model architecture was motivated by a compromise between simplicity (to allow fast experiments) and representativity (to ensure results are transferable). Same model as in~\cite{bso} was used: single-layer bidirectional LSTM encoder and single-layer LSTM decoder with multiplicative attention mechanism~\cite{attn}. Hidden dimension of LSTM cells is 256 and pretrained fasttext embeddings~\cite{fasttext} of size 300 were used for initialization. We used Adam~\cite{adam} with learning rate 0.001 when optimizing cross-entropy, and with learning rate 0.0001 when directly optimizing BLEU. We trained our models on dataset is IWSLT'14 dataset \cite{iwslt14} of German-English sentence pairs.

\begin{table}
	\caption{Experimental results for MT. Either differentiable loss function Cross-Entropy (CE) was used or we directly optimized the target BLEU metric (last line). }
    \label{results_mt}
    \centering
    \begin{tabular}{lcccc}
    \toprule
	& \multicolumn{2}{c}{CE loss} & \multicolumn{2}{c}{BLEU}\\
    \midrule
	& train & eval (best) & train & eval (best)\\
    \midrule
	Teacher-forcing (Figure~\ref{fig:scg}A, CE opt.) & $5.96 \pm 0.03$ & $6.44 \pm 0.08$ & $31.8 \pm 4.9$ & $19.6 \pm 3.6$\\
	\midrule
    Feed samples (Figure~\ref{fig:scg}B, CE opt.): & & & &\\
    naive gradient & $3.41 \pm 0.17$ & $4.20 \pm 0.10$ & $20.1 \pm 2.7$ & $9.8 \pm 1.3$\\
	full gradient, control variates & $3.45 \pm 0.23$ & $4.24 \pm 0.08$ & $12.4 \pm 1.7$ & $8.0 \pm 0.7$\\
	full gradient, Gumbel reparam. & $3.43 \pm 0.16$ & $4.22 \pm 0.07$ & $16.8 \pm 1.5$ & $8.7 \pm 0.3$\\

    \midrule
    full gradient (Figure~\ref{fig:scg}C, direct BLEU opt.) & - & - & $24.6 \pm 0.1$ & $22.0 \pm 0.2$\\
    \bottomrule
    \end{tabular}
\end{table}

Let us now discuss differences between training procedures for loss functions shown in Figure~\ref{fig:scg}. Teacher forcing case is trivial: there are no non-differentiable or stochastic nodes in the graph, so usual backpropagation algorithm works fine. 

If we introduce additional stochastic nodes in the graph, then in order to get an unbiased estimate for the gradient of loss function we need to backpropagate through stochastic nodes; the gradient consists of two terms in this case: the first one that flows through by all the paths in the graph that go through stochastic nodes, and the second one, that is just usual backpropagation term. We do not reproduce all the formulae here, they can be found in \cite{scg}. Hereinafter we refer to the sum of these two terms as "full" gradient, and just the second term is referred to as "naive" gradient.

For scheduled sampling, authors of \cite{scheduled_sampling} used naive gradient to train a model. It resulted in a biased estimate of gradient. One of the stimulating questions for the present work was to clarify how optimization process would change once we start using unbiased gradient estimates (but probably with larger variance). 

Introducing stochastic nodes allows us to optimize non-differentiable metrics, because in this case the expression for the full gradient does not include derivatives of cost nodes ("score function" estimator \cite{scg}). As an illustrative example, we do not need a model of environment when using policy gradient theorem \cite{pg} in RL just because of this. This is why one appeals to RL optimization procedures, once a non-differentiable metric occurs. As a price to pay, this estimator shows large variance \cite{var_red}. In our opinion, the whole RL machinery seems to be a bit redundant here, because we do not actually need to define of "agent", "environment" (or Markov decision process) here. We can just consider an arbitrary graph with stochastic nodes and optimize a corresponding loss function.

To summarize, there are three most common approaches to optimize loss functions represented as SCG:

\begin{enumerate}
\item Naive gradient (ignore paths through stochastic nodes): applicable to cases A and B on Figure~\ref{fig:scg}. Cannot be applied to case C, because loss function is non-differentiable.
\item Full gradient: universal, for case A in Figure~\ref{fig:scg} reduces to naive gradient.
\item Reparameterization trick: case B in Figure~\ref{fig:scg}. Same as naive gradient, cannot be applied to case C.
\end{enumerate}

The latter approach can introduce bias in gradient estimate (for example, if we use reparameterization for discrete distributions \cite{gumbel}), but usually greatly reduces variance. In this case we actually change computation graph, but do not change expected loss.

\section{Conclusions}
\label{conclusions}
SCG formalism provides a convenient framework for the analysis of machine learning architectures by showing assumptions made in corresponding approaches explicitly. Also, adding new stochastic nodes (i.e. hard attention mechanism \cite{hard_attn} instead of soft one) does not cause change in training paradigm from "supervised learning" to "reinforcement learning". SCG formalism allows to get rid of seemingly different approaches to the same task and review it from a common perspective.

Using sampling can be viewed as regularization technique: that is, comparing teacher-forcing and sampling approaches, we observed that the latter approach results in smaller margin between metrics on train and validation sets.

Using full gradient instead of naive one didn't provide any advancement in evaluation metrics in our experiments. This is likely due to high variance of gradient estimate. One way to mitigate it is to use better control variate (a promising example: \cite{rebar}), that remains a subject for future work.

\subsubsection*{Acknowledgments}
This work was supported by National Technology Initiative and PAO Sberbank project ID 0000000007417F630002.

\newpage
\bibliographystyle{apalike}
\bibliography{nips17opt}

\end{document}